\documentclass[10pt,journal]{IEEEtran}
\usepackage[pdftex]{graphicx}
\DeclareGraphicsExtensions{.pdf,.jpg,.png}
\usepackage{cite}
\usepackage[cmex10]{amsmath}
\usepackage{amssymb}
\usepackage{graphicx}
\usepackage{cases}
\usepackage{mdwmath}
\usepackage{mdwtab}
\usepackage{array}
\usepackage{url}
\usepackage[ruled,vlined]{algorithm2e}
\usepackage{booktabs}
\usepackage{threeparttable}
\usepackage{multirow}
\usepackage{color}
\usepackage{caption}

\def\ie{{\em i.e.}}
\def\eg{{\em e.g.}}
\def\etal{{\em et al.}}

\newcommand{\myPara}[1]{\vspace{.05in}\noindent\textbf{#1}}

\newcommand{\mc}[1]{\mathcal{#1}}
\newcommand{\mb}[1]{\mathbb{#1}}

\newcommand{\bm}[1]{\mbox{\boldmath{$#1$}}}

\graphicspath{{./fig/}}
\usepackage{color}
\begin{document}

\title{{Efficient Low-Resolution Face Recognition via Bridge Distillation}}
\author{Shiming Ge, \IEEEmembership{Senior Member,~IEEE,}
    Shengwei Zhao, Chenyu Li, Yu Zhang and
    Jia~Li, \IEEEmembership{Senior Member,~IEEE}
\thanks{S. Ge is with the Institute of Information Engineering, Chinese Academy of Sciences, Beijing, 100095, China. E-mail: geshiming@iie.ac.cn}
\thanks{S. Zhao and C. Li are with the Institute of Information Engineering, Chinese Academy of Sciences, Beijing 100095, China, and also with School of Cyber Security at University of Chinese Academy of Sciences, Beijing 100049, China. Email: \{zhaoshengwei,lichenyu\}@iie.ac.cn.}
\thanks{Y. Zhang is with SenseTime Group Limited, 100084, China.}
\thanks{J.~Li is with the State Key Laboratory of Virtual Reality Technology and Systems, School of Computer Science and Engineering, Beihang University. He is also with the Beijing Advanced Innovation Center for Big Data and Brain Computing, Beihang University, Beijing, 100191, China, and also with the Peng Cheng Laboratory, Shenzhen, 518055, China.}
\thanks{J. Li is the corresponding author. E-mail:~jiali@buaa.edu.cn}
}

\markboth{IEEE Transactions on Image Processing}%
{Ge \MakeLowercase{\textit{et al.}}: Bare Demo of IEEEtran.cls for Journals}

\maketitle

\begin{abstract}
  Face recognition in the wild is now advancing towards light-weight models, fast inference speed and resolution-adapted capability. In this paper, we propose a bridge distillation approach to turn a complex face model pretrained on private high-resolution faces into a light-weight one for low-resolution face recognition. In our approach, such a cross-dataset resolution-adapted knowledge transfer problem is solved via two-step distillation. In the first step, we conduct cross-dataset distillation to transfer the prior knowledge from private high-resolution faces to public high-resolution faces and generate compact and discriminative features. In the second step, the resolution-adapted distillation is conducted to further transfer the prior knowledge to synthetic low-resolution faces via multi-task learning. By learning low-resolution face representations and mimicking the adapted high-resolution knowledge, a light-weight student model can be constructed with high efficiency and promising accuracy in recognizing low-resolution faces. Experimental results show that the student model performs impressively in recognizing low-resolution faces with only 0.21M parameters and 0.057MB memory. Meanwhile, its speed reaches up to 14,705, ~934 and 763 faces per second on GPU, CPU and mobile phone, respectively.
\end{abstract}



\IEEEpeerreviewmaketitle

\section{Introduction}

\IEEEPARstart{A}{lthough} face recognition techniques become nearly mature for several real-world applications, they still have difficulties in handling low-resolution faces and being deployed on low-end devices \cite{Ge2019TIP}. These demands are very important for tasks like video surveillance and automatic driving. In general, most well-known face recognition models \cite{Schroff2015CVPR,liu2017cvpr,zheng2018ring,wang2018cosface,cao2018fg} are trained from massive high-resolution faces by using sophisticated architectures that contain huge parameters, making them uneconomical to deploy. Moreover, these models may be not suitable to directly apply on low-resolution scenarios due to the different distribution between high-resolution training faces (sometimes from private datasets) and low-resolution ones. An important reason is that the high-resolution facial details will be missing during the degeneration of the resolution, which the existing models largely depend on. An alternative way is to train a new model on massive low-resolution faces in target scenarios (\eg, surveillance faces in the wild). However, collecting and labeling such faces is very time and labor consuming. Moreover, directly training on low-resolution faces usually suffer from unsatisfactory accuracy \cite{wang2016cvpr}, since the reduction of image resolutions may lose some valuable knowledge which can be provided from some pretrained models \cite{Cheng2018AAAI}. Therefore, it is necessary to fully exploit the knowledge from massive high-resolution faces and pretrained models for facilitating low-resolution face recognition.
To recognize low-resolution faces, some feasible ideas based on hallucination or embedding are proposed to exploit high-resolution knowledge.

\begin{figure}[t]
  \begin{center}
  \includegraphics[width=1.0\linewidth]{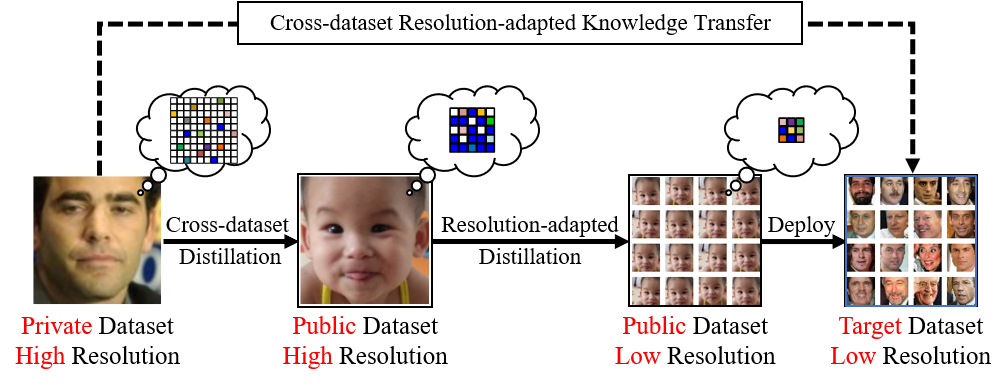}
  \end{center}
  \caption{Motivation of the bridge distillation. The direct knowledge transfer from private high-resolution faces to target low-resolution faces may be difficult. Therefore, we use public high-resolution and low-resolution faces as a bridge to step-wisely distil and compress the knowledge via cross-dataset distillation and resolution-adapted distillation. Note that the public low-resolution faces are generated from the public high-resolution faces to simulate the probable distribution of target low-resolution faces. }
  \label{fig:motivation}
\end{figure}

The ``hallucination'' idea is based on the fact that a person who is familiar with a high-resolution face can recognize the low-resolution counterpart. Several existing approaches propose to hallucinate the high-resolution faces before recognition by explicitly reconstructing details, such as \eg, with super-resolution \cite{jian2015simultaneous,yang2015recognition,yu2017cvpr,cheng2018accv}. Among them, the mapping between high-resolution to low-resolution faces is modeled by some carefully designed parametric functions (\eg~nonlinear Lagrangian \cite{kolouri2015transport}, SVD \cite{jian2015simultaneous} and sparse representation \cite{yang2015recognition}). During inference, parametric coefficients that best fit the given low-resolution faces are computed and adopted to recover the missing details to make recognition easier. These approaches generally can achieve good recognition accuracy, while the additional reconstruction often brings in computational burden and slows down the recognition speed.

Different from the hallucination-based approaches, the ``embedding'' idea applies an implicit scheme to directly project low-resolution faces into an embedding space that mimics high-resolution behaviors. In this way, only the high-resolution features are encoded into the model without explicit reconstruction. For example, Biswas \etal \cite{biswas2012multidimensional} embedded low-resolution facial images into an Euclidean space, in which image distances well approximate the dissimilarities among high-resolution images. Ren \etal \cite{ren2012coupled} projected face images of different resolutions into a unified space for coupled matching. A shared characteristic of such embedding models is to encode informative high-resolution details into low-resolution features via cross-resolution analysis. They are generally model-specific and not designed to mimic an existing high-resolution model whose pre-learned knowledge is not fully adopted. However, the pre-learned knowledge often contains rich high-resolution details and can guide low-resolution face recognition if being properly adapted and transferred.

Knowledge distillation is an efficient way to transfer knowledge via the teacher-student framework \cite{Hinton2014NIPSW,Romero2015ICLR,Luo2016AAAI,lopezpaz2016iclr,radosavovic2018cvpr,Phuong2019icml}. In the framework, the teacher is usually a strong yet complex model that performs well on its private dataset, while a much simpler student is learned to mimic the teacher's behavior, leading to maximal performance preservation and speed improvement. The key of knowledge distillation is the trade-off between speed and performance, and such a technique provides an opportunity to convert many complex models into simple ones for practical deployment. However, most of existing distillation approaches assume that teacher and student training are restricted on the same dataset or the same resolution, which is not suitable in many real scenarios where a well-pretrained teacher on an existing dataset would like to be reused to supervise model training on a new dataset. In \cite{Ge2019TIP}, Ge \etal proposed a selective knowledge distillation approach to transfer the most informative knowledge from pre-trained high-resolution teacher to a lightweight low-resolution student by solving a sparse graph problem, which actually performed cross-resolution distillation so that the accuracy of low-resolution face recognition can be improved. However, the selected knowledge may not be optimal in adapting on the training faces.

In summary, transferring the knowledge from high-resolution to low-resolution models is helpful and can avoid the computationally-intensive reconstruction. Thus, a learning framework to help recognize low-resolution faces should be able to effectively transfer informative high-resolution knowledge in a principle manner. That is to say, It need to actually solve two subproblems: \textbf{what knowledge should be transferred} from the high-resolution models and \textbf{how to perform such transfer}. In this way, the challenges in low-resolution face recognition and knowledge distillation are simultaneously addressed with a single framework.

Inspired by that, as shown in Fig. \ref{fig:motivation}, we propose a novel bridge distillation approach that can convert existing high-resolution models pretrained on their private datasets into a much simpler one for low-resolution face recognition on target dataset. In our approach, public high-resolution faces and their resolution-degraded versions are used as a bridge to compress a complex high-resolution teacher model to a much simpler low-resolution student model via two step distillation. The first cross-dataset distillation adapts the pretrained knowledge from private to public high-resolution faces. It learns a feature mapping that preserves both the discriminative capability on public high-resolution faces as well as the detailed face patterns encoded in the original private knowledge. Then, the second resolution-adapted distillation learns a student model in a multi-task fashion to jointly mimic the adapted high-resolution knowledge and recognize the public low-resolution faces, which are synthesized to simulate the probable distribution of the target low-resolution faces. In this way, the student model only needs to be aware of the high-resolution details that are still discriminative on low-resolution faces regardless of the others, resulting into compact knowledge transfer.

The contributions are summarized as follows: 1)~we propose a bridge distillation framework that is able to convert high-resolution face models to much simpler low-resolution ones with greatly reduced computational and memory cost as well as minimal performance drop; 2)~we propose cross-dataset distillation, which adapts the pre-learned knowledge from private to public high-resolution faces that preserves the compact and discriminative high-resolution details; 3)~comprehensive experiments are conducted and show that the student models achieve comparable accuracy with the state-of-the-art high-resolution face models, but with extremely low memory cost and fast inference speed.

\section{Related Works}
In this section, we first briefly review the development of low-resolution face recognition models, then introduce knowledge distillation directions which are tightly correlated with the proposed approach in greater details.

\subsection{Low-Resolution Face Recognition}
Recently, deep learning approaches have motivated many strong face recognition models, \eg \cite{Sun2014NIPS,liu2016icml,Zhang2017RangeLoss,cao2018fg,wang2018cosface}. However, the performance of these models may drop sharply if applied to low-resolution faces. An important reason is that the high-resolution facial details will be missing during the degeneration of the resolution, which the existing models largely depend on. To address this problem, several recent works propose to adopt two ideas to handle low-resolution recognition: \textit{hallucination} and \textit{embedding}, which reconstructs the high-resolution details explicitly or implicitly during inference.

In the hallucination category, high-resolution facial details are explicitly inferred and then utilized in the recognition process. For example, Kolouri \etal \cite{kolouri2015transport} proposed to fit the low-resolution faces to a non-linear Lagrangian model, which explicitly considers high-resolution facial appearance. Jian \etal \cite{jian2015simultaneous} and Yang \etal \cite{yang2015recognition} instead adopted SVD and sparse representation to jointly performing face hallucination and recognition. Several works \cite{ledig2016photo,Zhang2018ECCV} were able to generate highly realistic high-resolution face images from low-resolution input. Cheng \etal \cite{cheng2018accv} introduced a complement super-
resolution and identity joint deep learning method with a unified end-to-end network architecture to address low-resolution face recognition. Although hallucination is a direct way to address low-resolution face recognition, it is usually computationally intensive as it introduces face reconstruction as a necessary pipeline. Li \etal \cite{li2019low} introduced a GAN pre-training approach and fully convolutional architecture to improve face re-identification and employed supervised discriminative learning to explore low-resolution face identification. Recent face recognition models proposed designing effective loss functions \cite{liu2016icml,wang2018cosface,liu2017nips,wang2018additive,liu2018nips,deng2019cvpr} for feature learning, that can facilitate the hallucination-based approaches.

In contrary, the embedding category implicitly encodes high-resolution features in low-resolution computation to avoid high-resolution reconstruction. This inspires several embedding-based approaches, which project both high-resolution and low-resolution faces into a unified feature space so that they can be directly matched \cite{jiang2016cdmma,wang2016pose,haghighat2017low}. This paradigm can be implemented by jointly transforming the features of high-resolution and low-resolution faces into a unified space using multidimensional scaling \cite{mudunuri2016low}, joint sparse coding \cite{shekhar2017synthesis} or cross-resolution feature extraction and fusion \cite{li2015multi,pong2014multi}. Recently, deep learning were adopted for low-resolution face recognition in \cite{wang2016cvpr} and \cite{herrmann2016low}. However, most of these works focus on joint high- and low-resolution training, while the distillation of high-resolution models for low-resolution tasks is still an open problem.

From these works, we find transferring the knowledge from high-resolution to low-resolution models is helpful and can avoid the computationally-intensive face reconstruction. Thus, we need to actually solve two subproblems: what knowledge should be transferred from the high-resolution models and how to perform such transfer. Therefore, we also briefly review knowledge distillation studies.

\subsection{Knowledge Distillation and Transfer}
Knowledge distillation \cite{Hinton2014NIPSW} is a useful way that utilizes a strong model to supervise a weak one, so that the weaker model can be improved on the target task and carry out domain adaptation \cite{rozantsev2018cvpr}. In particular, the teacher-student framework is actively studied \cite{Romero2015ICLR,Luo2016AAAI,chen2017nips,yim2017gift,radosavovic2018cvpr}, where the teacher model is usually a strong yet complex model that performs well on its private dataset. Knowledge distillation will learn a new and much simpler student model to mimic the teacher's behavior under some constraints, leading to maximal performance preservation and speed improvement. In this manner, the learned student model can recover some missing knowledge that may not be captured if it is trained independently. Among them, Luo \etal \cite{Luo2016AAAI} proposed to distil a large teacher model to train a compact student network. In their approach, the most relevant neurons for face recognition were selected at the higher hidden layers for knowledge transfer. Su and Maji \cite{su2017bmvc} utilized cross quality distillation to learn models for recognizing low-resolution images. Lopez-Paz \etal \cite{lopezpaz2016iclr} proposed the general distillation framework to combine distillation and learning with privileged information. There are some approaches that focus on the tasks of continue or incremental learning. In \cite{rebuffi2017cvpr}, Rebuffi \etal introduced iCaRL, a class-incremental training strategy that allows learning strong classifiers and a data representation simultaneously by presenting a small number of classes and progressively adding new classes.
In \cite{li2018pami}, Li and Hoiem proposed Learning without Forgetting method that uses only new task data to train the network while preserving the original capabilities. Typically, these continue or incremental learning approaches are different from those feature extraction and fine-tuning adaption techniques as well as multitask learning that uses original task data.

\begin{figure*}[t]
	\begin{center}
		\includegraphics[width=1.0\linewidth]{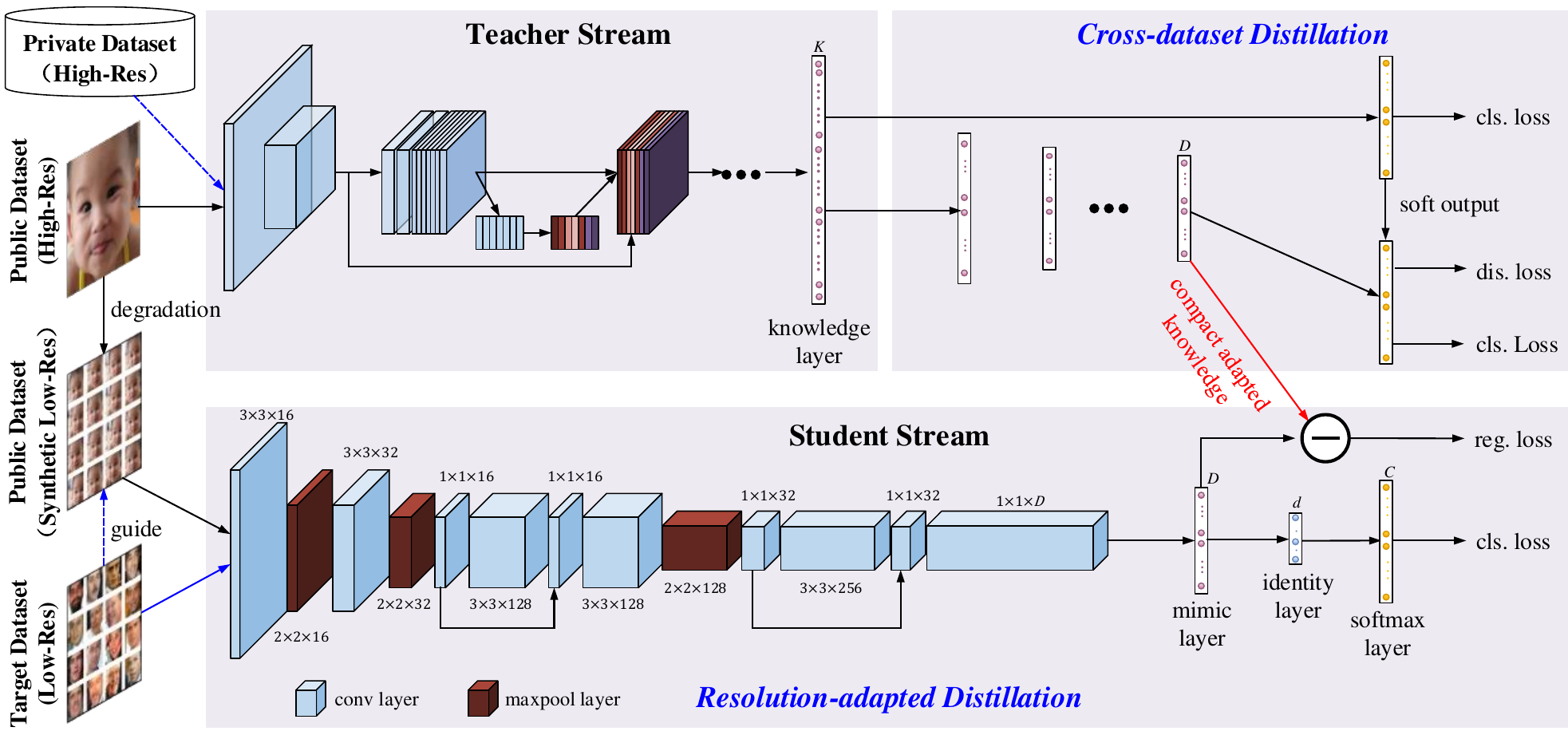}
	\end{center}
	\caption{The bridge distillation framework, which consists of a teacher stream and a student stream. 1) The teacher stream is first pre-trained on high-resolution private faces, which extracts the learned knowledge about informative facial details. Then, cross-dataset distillation adapts the learned knowledge to the high-resolution public faces so as to preserve compact and discriminative features. 2) The student stream is trained on low-resolution face recognition via resolution-adapted distillation by jointly performing two tasks: feature regression to mimic the adapted high-resolution knowledge, and face classification on low-resolution public faces. Thus, the resulting student models could be deployed to recognize low-resolution target faces.}
	\label{fig:framework}
\end{figure*}

To sum up, the core of knowledge distillation is the trade-off between speed and performance, and such a technique provides an opportunity to convert complex models into simple ones that can be deployed in the wild. However, most of these works assume that teacher and student training are restricted on the same dataset or the same resolution. In many real scenarios, we would like to reuse a well-pretrained teacher model on an existing dataset to supervise model training on a novel dataset, which cannot be directly handled by existing approaches. Moreover, we attempt to use such a learning framework to help recognize low-resolution faces, since informative high-resolution knowledge can be effectively transferred in a principle manner.
In this way, the challenges in low-resolution face recognition and knowledge distillation are simultaneously addressed with a single framework.

\section{The Proposed Approach}
\subsection{Framework}
Our bridge distillation approach is an intuitional and general framework (see Fig. \ref{fig:framework}) to convert an existing high-resolution model to a simpler one, which is expected to work well for low-resolution recognition. The framework basically follows the teacher-student framework \cite{Hinton2014NIPSW} but with two-step distillation. A complex teacher model is first trained on the high-resolution face recognition task, then distilled to a simpler student model for low-resolution scenario. The framework refers to the following notations.

\textbf{Domains.}~The framework involves three domains: 1) \textit{private domain} refers to an external high-resolution private face dataset $\mb{I}_P$ for training the teacher model, which is usually large and unnecessary to be visible to the framework, 2) \textit{public domain} is the public face dataset $\mb{I}_S$ and used as a bridge to learn a simple student model by adapting and mimicking the teacher's knowledge, and 3) \textit{target domain} refers to the deployment scenario for recognizing unseen low-resolution faces $\mb{I}_T$ (\eg, surveillance faces). Without loss of generality, we assume the face distribution in the target domain is observable but the labels are not available.

\textbf{Teacher model.} In the proposed framework, the teacher model is assumed to be \textit{off-the-shelf}, meaning that it is pretrained on $\mb{I}_P$. In general, the pretrained teacher model encodes rich and general knowledge of high-resolution face recognition. When applied to a novel dataset (\eg, $\mb{I}_T$), it is thus desirable to transfer the pre-learned high-resolution knowledge across datasets rather than retraining the model from scratch. Note that it differs from many teacher-student frameworks \cite{Hinton2014NIPSW,Romero2015ICLR,Luo2016AAAI,chen2017nips,yim2017gift,radosavovic2018cvpr}, where training is restricted on the private dataset or the same resolution or available target dataset. In this work, we assume that the teacher model $\mc{M}_t$ is in the form $\mc{M}_t = \left( \mc{F}_t, \mc{S}_t \right)$, composed by a feature extraction backend $\mc{F}_t$ and a softmax classification layer $\mc{S}_t$. This is a widely applied architecture which most face classification models (\eg \cite{cao2018fg,wang2018cosface}) conform. Thus, given an input face image $\mc{I}$, high-level features are first computed with $\bm{\mathrm{f}}_t \left(\mc{I}\right)=\mc{F}_t \left( \mc{I}; \bm{\mathrm{w}}^f_t \right)$, which are processed via $\bm{\mathrm{s}}_t(\mc{I}) = \mc{S}_t \left(\bm{\mathrm{f}}_t\left( \mc{I} \right); \bm{\mathrm{w}}^s_t \right)$ to obtain classification scores. Here, $\bm{\mathrm{w}}_t = \left[\bm{\mathrm{w}}^f_t; \bm{\mathrm{w}}^s_t \right]$ denotes the model parameters. The design of $\mc{F}_t$ and $\mb{I}_T$ can be blind to the proposed approach.

\textbf{Student model.} The student model has much simpler design than that of the teacher. By training the student to mimic the teacher's behavior on the public dataset $\mb{I}_S$, the model's complexity is largely reduced. In practical settings, our aim is to obtain an optimized student model $\mc{M}_s$ that works well on the low-resolution target dataset $\mb{I}_T$. Usually $\mb{I}_S$, $\mb{I}_T$ and $\mb{I}_P$ do not share identities, \ie~$\mb{I}_S \cap \mb{I}_T=\phi$ and $\mb{I}_S \cap \mb{I}_P=\phi$. The difference in resolution also exists between private and target datasets. Thus, the problem is how to properly transfer the high-resolution knowledge well-learned from $\mb{I}_P$ to facilitate low-resolution recognition on $\mb{I}_T$ by using $\mb{I}_S$. Such transfer needs to address the distribution and resolution differences simultaneously.

\textbf{Formulation.} Given the above settings, low-recognition face recognition can be formulated as an open-set domain adaptation problem \cite{busto2017open} in cross-resolution context, where the goal is to achieve effective knowledge transfer from the discriminative high-resolution teacher $\mc{M}_t \left(\mc{I}; \bm{\mathrm{w}}_t \right)$ to a simpler low-resolution student $\mc{M}_s \left(\mc{I}'; \bm{\mathrm{w}}_s \right)$, with huge domain shift between the private dataset $\mb{I}_p$ and the target dataset $\mb{I}_t$. Here, the two datasets usually do not share identities, leading to the difference in characteristics distribution. Moreover, the extra resolution differences further adds to the domain shift. Therefore, the knowledge transfer needs to address the distribution and resolution differences simultaneously.
To this end, we introduce public source faces $\mb{I}_s$ as bridging domain and propose bridge distillation to perform transfer via two step distillations: 1) \textit{cross-dataset distillation} adapts the pre-learned knowledge in high-resolution faces from private domain to public domain and distils it to compact features, and 2) \textit{resolution-adapted distillation} transfers the knowledge from high-resolution public faces to their low-resolution versions by taking the adapted features as supervision signals to train the student model. In this context, $\mb{I}_s$ could be easily achieved from many public benchmarks, meanwhile many off-the-shelf face recognition models are available and can serve as teachers. Therefore, our approach provides an economical way to learn an efficient model for facilitating low-resolution face recognition.


\subsection{Cross-dataset Distillation}
Cross-dataset distillation addresses the subproblem that \textbf{what knowledge should be transferred} from the high-resolution models. It is used to compress and adapt the teacher's knowledge learned from $\mb{I}_p$ to $\mb{I}_s$ with an adaptation process which should take two considerations into account. First, it should preserve high-resolution knowledge learned from $\mb{I}_p$, while selectively enhancing them to be discriminative on $\mb{I}_s$ so that the correct high-resolution knowledge could be extracted from $\mb{I}_s$ with the teacher. Second, the adapted features should be compact so that the student can mimic them with extremely limited resources.

Previous works show that directly training on low-resolution face images usually suffer from unsatisfactory recognition accuracy \cite{wang2016cvpr}. The rationale of this degeneration is that important face details gradually lose when image resolution reduces. To mitigate this problem, high-resolution face knowledge should be encoded into the trained model, so that detail reconstruction can be implicitly performed during the recognition process.

In our setting, the teacher model $\mc{M}_t \left(\mc{I}; \bm{\mathrm{w}}_t \right)$ learned on the rich high-resolution dataset $\mb{I}_P$ is usually strong for recognizing high-resolution faces from its own dataset. Here $\mc{I}$ represents the input high-resolution face image, and $\bm{\mathrm{w}}_t$ is the concatenation of model parameters. We expect the teacher's knowledge can be transferred to the student model $\mc{M}_s \left(\mc{I}'; \bm{\mathrm{w}}_s \right)$, whose input is a low-resolution face image $\mc{I}'$ that comes from separate datasets $\mb{I}_S$ or $\mb{I}_T$. However, we assume that $\mb{I}_S$ also contains high-resolution faces, so that the teacher network can learn the high-resolution details.
For better clearness, we use the notation $\mc{I}$ and $\mc{I}'$ to represent a high-resolution and low-resolution face image, respectively. Given that $\mb{I}_P$ and $\mb{I}_S$ can have distinct distributions, we propose to first adapt the teacher's knowledge learned from $\mb{I}_P$ to $\mb{I}_S$. The adaptation process should take two considerations into account. First, it should preserve the high-resolution knowledge learned from $\mb{I}_P$, while selectively enhancing them to be discriminative on the public dataset $\mb{I}_S$. Second, the adapted features should be compact, so that the student model can mimic them with extremely limited computational resources.

To implement this idea, we instantiate the adaptation function as a small sub-network $\mc{F}_a \left(\bm{\mathrm{f}}_t \left( \mc{I} \right); \bm{\mathrm{w}}^f_a \right)$ that maps the high-dimensional features $\bm{\mathrm{f}}_t \left( \mc{I} \right)$ into a reduced feature space. By plugging this adaptation module into the teacher model, optimal feature mapping can be learned in a data-driven fashion. Given the adapted features $\bm{\mathrm{f}}_a \left( \mc{I} \right) = \mc{F}_a \left(\bm{\mathrm{f}}_t \left( \mc{I} \right); \bm{\mathrm{w}}^f_a \right)$, we assume that a simple softmax classifier $\mc{S}_a \left( \bm{\mathrm{f}}_a \left( \mc{I} \right); \bm{\mathrm{w}}^s_a \right)$ recognizes the faces in the public dataset $\mb{I}_S$ well. Denote $\mc{M}_a = \left( \mc{F}_a, \mc{S}_a \right)$ as the full adaptation module, where $\bm{\mathrm{w}}_a = \left[ \bm{\mathrm{w}}^f_a; \bm{\mathrm{w}}^s_a \right]$ are its parameters to be learned. We propose to train the adaptation module with the following objective, which can be deemed as a variant of knowledge distillation \cite{Hinton2014NIPSW,lopezpaz2016iclr},
\begin{equation}
\min_{\bm{\mathrm{w}}_a} \mc{C} \left(\bm{\mathrm{w}}_a, \mb{I}_S \right) + \lambda \mc{D} \left(\bm{\mathrm{w}}_a, \mb{I}_S \right).
\label{eq:self-distillation}
\end{equation}
The proposed objective is a balanced sum of a classification loss $\mc{C}$ and a distillation loss $\mc{D}$, with $\lambda$ as the balancing weight. For classification on the high-resolution public faces, we have
\begin{equation}
\mc{C} \left(\bm{\mathrm{w}}_a, \mb{I}_S \right) = \sum_{\mathcal{I} \in \mathbb{I}_S} \ell \left(\mc{M}_a \left(\bm{\mathrm{f}}_t \left( \mc{I} \right); \bm{\mathrm{w}}_a \right), \bm{\mathrm{y}} \left(\mc{I}\right)\right),
\label{eq:classification_loss}
\end{equation}
where we adopt the widely used cross-entropy classification loss $\ell \left( \cdot, \cdot \right)$, and $\bm{\mathrm{y}} \left( \mc{I} \right)$ is the one-hot groundtruth identity for input image $\mc{I}$. The distillation loss $\mc{D}$ enforces the adapted features to mimic the original features' behavior on face classification. To this end, we first fine-tune the teacher's softmax layer on $\mb{I}_S$, obtaining $\hat{\mc{S}}_t \left(\bm{\mathrm{f}}_t \left(\mc{I}\right); \hat{\bm{\mathrm{w}}}^s_t \right)$, where $\hat{\bm{\mathrm{w}}}^s_t$ denotes the retrained layer weights. One can think of $\hat{\mc{S}}_t$ as a feature selector that preserves the discriminative components of the originally learned knowledge for recognizing faces in $\mb{I}_S$. Then the distillation loss is given as:
\begin{equation}
\mc{D} \left(\bm{\mathrm{w}}_a, \mb{I}_S \right) = \sum_{\mathcal{I} \in \mathbb{I}_S} \ell \left(\mc{M}_a \left(\bm{\mathrm{f}}_t \left( \mc{I} \right); \bm{\mathrm{w}}_a \right), \hat{\bm{\mathrm{s}}} \left(\mc{I}\right)\right),
\label{eq:distillation_loss}
\end{equation}
where $\hat{\bm{\mathrm{s}}}\left(\mc{I}\right)=\hat{\mc{S}}_t \left(\bm{\mathrm{f}}_t \left(\mc{I}\right); \hat{\bm{\mathrm{w}}^s_t} \right)/T$ and $T$ is the temperature \cite{Hinton2014NIPSW} for softening the softmax outputs. Eq.\ref{eq:distillation_loss} uses cross-entropy to approximate the softened teacher's softmax features, which ensures the adaptation process bringing in the additional teacher's knowledge learned from $\mb{I}_P$. In addition to the recognition capacity regarding $\mb{I}_S$ given by Eq.\ref{eq:classification_loss}, the knowledge of the teacher network is retained. As a result, the adapted features mimic the impact of the discriminative selection of the original high-resolution features, but with greatly reduced dimensions. After training the adaptation module $\mc{M}_a$, we discard the softmax layer and only retain the adapted and reduced features $\bm{\mathrm{f}}_a \left(\mc{I}\right)=\mc{F}_a \left(\bm{\mathrm{f}}_t \left( \mc{I} \right); \bm{\mathrm{w}}^f_a \right)$ as supervision for training the student model.

\textbf{Difference from Other Distillation Approaches.} Different from classic distillation approaches\cite{Hinton2014NIPSW,lopezpaz2016iclr}, the proposed bridge distillation approach goes through two steps of distillations: the first step adapts the pretrained complex model to the public dataset, and the second step learns to mimic it with a simpler model. As the first step distils the model itself cross private and public datasets, we call the proposed algorithm \textit{cross-dataset distillation}. Thus, the teacher need not to be fully trained on the public dataset. Directly retraining the teacher on the public dataset not only costs a lot of time, but may also overfit to the dataset and lose the previously learned knowledge. By contrast, the proposed adaptation approach preserves such information, and is much faster to train.

\textbf{Difference from Other Learning Approaches.} Recent continue or incremental learning approaches \cite{rebuffi2017cvpr,li2018pami} focus on using new task data to train the network while preserving the knowledge learned from original task data. By contrast, our setting of \textit{cross-dataset distillation} mainly performs knowledge adaptation rather than knowledge preservation such that the teacher capacity can be transferred to public domain, which can facilitate the knowledge alignment between high-resolution and low-resolution instances. In this way, the capacity on high-resolution recognition can be preserved while the adapted knowledge into public high-resolution dataset can avoid being contaminated, leading to effective knowledge transfer. We also note that our \textit{cross-dataset distillation} combines knowledge adaptation and knowledge transfer together, which is carried out by fine-tuning and transfer learning.

\subsection{Resolution-adapted Distillation}
Resolution-adapted distillation addresses the subproblem that \textbf{how to perform knowledge transfer} from high-resolution faces to low-resolution ones. Given the high-resolution knowledge adapted to the public dataset, we ask the student to approximate them during inference. Since the capacity of the student is weak, the feature layer that mimics the adapted knowledge should be sufficiently deep. Empirically, we find that the mimicking layer is best inserted before the identity layer for softmax classification, as shown in Fig. \ref{fig:framework}.

Based on this design, we divide $\mc{M}_s$ to $\mc{M}^l_s$ and $\mc{M}^h_s$, \ie~the lower and the higher part, and its parameters $\bm{\mathrm{w}}_s$ to $\bm{\mathrm{w}}^l_s$ and $\bm{\mathrm{w}}^h_s$ accordingly. The lower part $\mc{M}^l_s$ corresponds to the main feature branch till the mimicking layer, and $\mc{M}^h_s$ consist of the rest layers. The student model is trained using the following objective
\begin{equation}
\min_{\bm{\mathrm{w}}^h_s, \bm{\mathrm{w}}^l_s} \hat{\mc{C}} \left(\bm{\mathrm{w}}_s, \mb{I}_S \right) + \mc{R} \left(\bm{\mathrm{w}}^l_s, \mb{I}_S \right),
\end{equation}
where face classification and feature regression are combined in a unified multi-task learning task. For the classification loss $\hat{\mc{C}}$ we still adopt the cross-entropy, but this time perform training on the degraded low-resolution versions of face images from the public dataset $\mb{I}_S$:
\begin{equation}
\hat{\mc{C}} \left(\bm{\mathrm{w}}_s, \mb{I}_S \right) = \sum_{\mathcal{I} \in \mathbb{I}_S} \sum_{\mathcal{I}' \in \mathbb{D} \left( \mc{I} \right)} \ell \left(\mc{M}_s \left(\mc{I}'; \bm{\mathrm{w}}_s \right), \bm{\mathrm{y}} \left(\mc{I}\right)\right),
\label{eq:face_classify}
\end{equation}
where $\mb{D} \left( \mc{I} \right)$ denotes the set of images degraded from $\mc{I}$.

The regression loss $\mc{R}$ is defined as
\begin{equation}
\mc{R} \left( \bm{\mathrm{w}}_s, \mb{I}_S \right) = \sum_{\mc{I} \in \mb{I}_S} \sum_{\mc{I}' \in \mb{D} \left( \mc{I} \right)} \| \mc{M}^l_s \left(\mc{I}'; \bm{\mathrm{w}}_s^l \right) - \mc{F}_a \left(\bm{\mathrm{f}}_t \left(\mc{I}\right); \bm{\mathrm{w}}_a \right)\|^2.
\end{equation}
After training, the teacher model and the adaptation module are discarded. During inference, the student model takes as input a low-resolution face image and outputs its classified identity features or labels, depending on the task.

\subsection{Implementation Details}
The proposed framework is designed to be flexible, so that in principle the teacher model can take form of any models or their ensemble as long as they end up with a softmax layer for face classification. Note that many state-of-the-art models \cite{Parkhi2015BMVC,Schroff2015CVPR,cao2018fg,wang2018cosface} meet this assumption. In this work, we adopt two most recent architecture VGGFace2 \cite{cao2018fg}, CosFace \cite{wang2018cosface} with a $112 \times 112$ input resolution and a $1024$ embedding feature dimension, and their ensemble in the teacher model for example. VGGFace2 is pretrained on massive high-resolution face images from the VGGFace2 dataset \cite{cao2018fg} and works at resolution $224 \times 224$, while CosFace is pretrained on CASIA-WebFace dataset \cite{yi2014webface} and a large-scale private dataset. No retraining or fine-tuning is performed on these datasets during evaluation. Also, we assume that their datasets are private, \ie, they cannot be accessed by the proposed approach.

For the student model, we design a light-weight architecture similar to the ones proposed in \cite{Redmon2017CVPR,He2016CVPR}. As shown in Fig. \ref{fig:framework}, it takes as input a low-resolution face image. We train the student model using various resolutions of $\textbf{p} \times \textbf{p}$, where $\textbf{p}=\{96,64,32,16\}$. The architecture has ten convolutional layers, three max pooling layers and three fully connected layers, interleaved by ReLU non-linearities. Several $1 \times 1$ convolution layers are intersected between $3 \times 3$ ones to save storage and improve inference speed. Two skip connections are established to enhance the information flow. Global average pooling is used to make final prediction so that the architecture can handle arbitrary resolutions. With this architecture, the amount of parameters is only $0.21$M, which is only $0.81\%$ or $0.57\%$ of the teacher's size ($26$M for VGGFace2 or $37$M for CosFace).

Our adaptation module takes the features before the teacher model's softmax layer as input. It has two fully connected layers with $512$ and $128$ units, respectively. More complex architecture can be used at the cost of additional training time. All the weights in the student model and the adaptation module are initialized via Xavier's method.
During training, the student model is first pretrained on the low-resolution faces in $\mb{I}_S$, then fine-tuned with the supervision adapted from the high-resolution knowledge extracted by the teacher model. Training images from $\mb{I}_S$ are downsampled by different factors to simulate degeneration at various levels. Batch normalization is performed to accelerate convergence speed. We use stochastic gradient descent to train the student models. In all the experiments, the batch size and learning rate are set as $256$ and $0.001$, respectively.

\section{Experiments}

\begin{figure}[t]
	\centering{\includegraphics[width=1.0\linewidth]{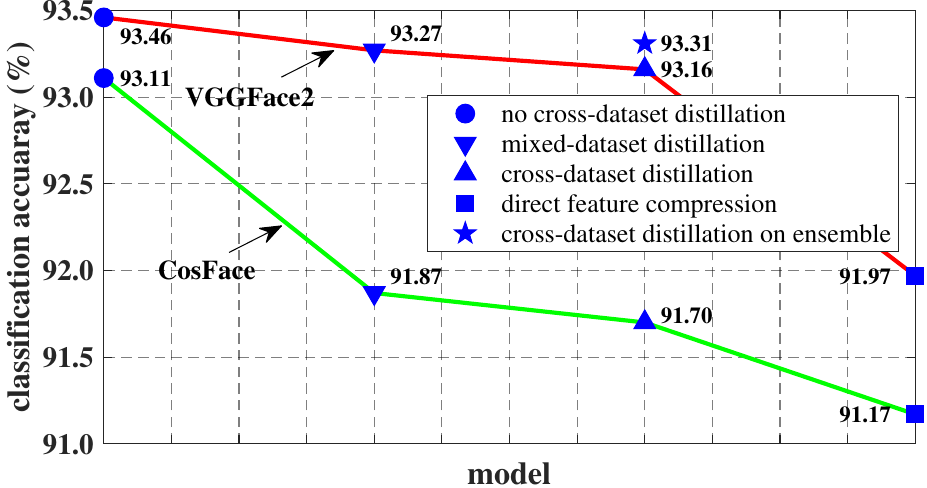}}
	\caption{The performance of the adapted models with and without cross-dataset distillation on UMDFaces.}
	\label{fig:res-umdfaces}
\end{figure}

\begin{figure*}[t]
	\centering{\includegraphics[width=1.0\linewidth]{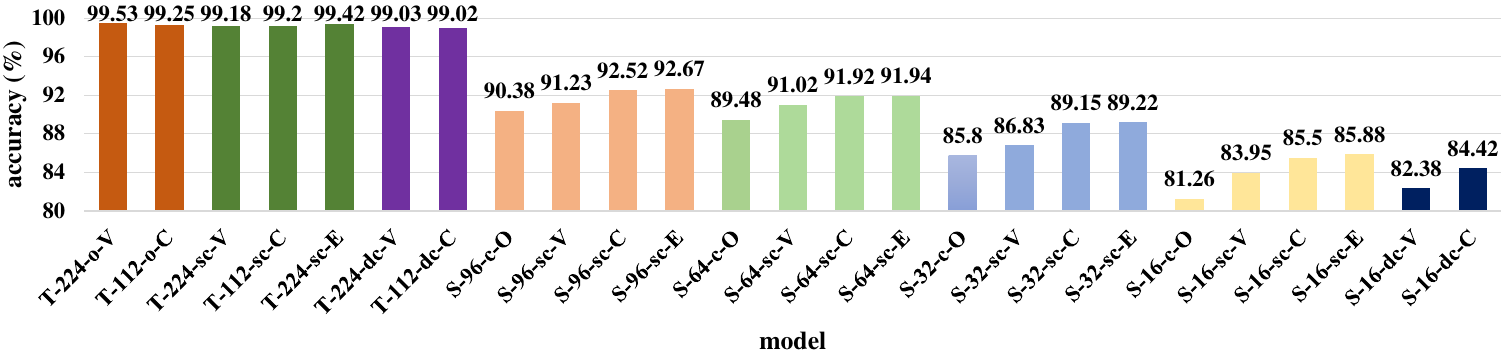}}
	\caption{The face verification performance of various teacher and student models on LFW.}
	\label{fig:res-lfw}
\end{figure*}

To validate the proposed approach, we conduct extensive experiments on three challenging public benchmarks: UMDFaces \cite{Bansal2016UMDFaces} dataset, LFW (Labeled Faces in the Wild) \cite{Learned2016Labeled} dataset and UCCS (UnConstrained College Students) \cite{Sapkota2013BTAS} dataset, for different evaluation tasks.

UMDFaces serves as the public dataset $\mb{I}_S$, which contains $367,888$ images in $8,419$ subjects. To generate high-resolution public images, faces are normalized into $224\times224$ and $112\times112$ sizes for learning the VGGFace2 and CosFace adaptation module. Low-resolution public face images are achieved by randomly perturbing the localized facial landmarks for $16$ times, normalizing to various resolutions of $\textbf{p}\times\textbf{p}$ and degrading to approximate the distribution of target faces (\eg, blurring, changing illumination, etc). Among all these high- and low-resolution images, $80\%$ of them are randomly selected for training and the rest for evaluating. 

LFW is used for target dataset, where $6,000$ pairs (including $3,000$ positive and $3,000$ negative pairs) are selected to evaluate various models on face verification task. To this end, the images are resized according to the model input. We extract the features from the mimicking layer and the identity layer of the student model for each input pair. Cosine similarity is computed for verification using a threshold. This experiment aims to show that better supervision from the high-resolution models can help generate better features in supervising the training of student models. 

UCCS contains $16,149$ images in $1,732$ subjects. It is a very difficult dataset with various levels of challenges, including blurred image, occluded appearance and bad illumination. The identities in the training and testing datasets are exclusive. It is widely used to benchmark face recognition models in unconstrained scenario. Thus, on this dataset we compare the proposed approach with the state-of-the-art models on face identification task, following the standard top-K error metric \cite{Krizhevsky2012NIPS}.

Generally, the high-resolution face images are available from many public datasets (e.g., UMDFaces), which fulfills the training of student models by transferring knowledge or recognition ability from high-resolution public faces to low-resolution ones synthesized according to the distribution of target faces. As a result, the trained student models could recognize low-resolution face images in the wild (even the high-resolution images are not available in this case, such as surveillance face images in UCCS). All the experiments across this paper are conducted using a NVIDIA K80 GPU, a single-core 2.6HZ Intel CPU and a mobile phone with a Qualcomm Snapdragon 821 processor implemented with TensorFlow framework \cite{Abadi2016OSDI}.
\begin{figure}[t]
  \begin{center}
     \includegraphics[width=1.0\linewidth]{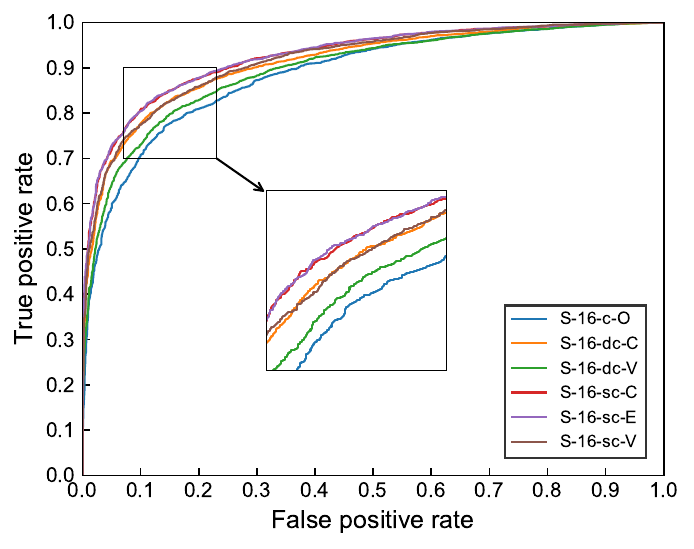}
  \end{center}
     \caption{ROC curves of various models on LFW.}
  \label{fig:roc-lfw}
  \end{figure}

\subsection{Evaluating Adaptation on UMDFaces} %
In the first experiment, we aim to look into the improvement brought by the proposed adaptation module via cross-dataset distillation. To this end, we evaluate four cases: 1) No cross-dataset distillation. In this case, the pretrained $2,048$ or $1,024$ dimensional features extracted by the VGGFace2 or CosFace teacher model are directly used to supervise the student model. 2) Direct feature compression without distillation. This setting adapts the pre-learned teacher's knowledge to the low-resolution setting with classification loss only, but excludes the distillation loss. 3) The proposed cross-dataset distillation, which equals to the formula \eqref{eq:self-distillation} that combines both losses. 4) Mixed-dataset distillation. In this case, we assume that the private high-resolution dataset can be accessible and used for training. In our experiment, we random select the training instances in 1,000 subjects from private high-resolution dataset (VGGFace2 dataset for VGGFace2 model or CASIA-WebFace dataset for CosFace model) and combine it with UMDFaces training set as a mixed training dataset, and then train a larger 9,419-way classifier to adapt the teacher knowledge into public high-resolution dataset.

In the last three cases, features are reduced to $128$ dimensions by adaptation. Fig.\ref{fig:res-umdfaces} shows the recognition accuracy of these four cases on UMDFaces, indicating that feature compression via cross-dataset distillation has a better performance than direct compression. Moreover, performance only slightly drops from case 1 to 3, especially cross-dataset distillation on the ensemble of two teachers has a $0.15\%$ drop against the supervised student model by VGGFace2 without cross-dataset distillation, while features are greatly reduced $128$ dimensions, thus greatly compressing the redundancy and saving resource consumption.
In addition, the trained classifier for mixed-dataset distillation achieves the accuracies of 91.87\% and 93.27\% with VGGFace2 and CosFace models, respectively. It shows that our cross-dataset distillation only has a very small drop in accuracy without ensemble, while achieving an accuracy improvement with ensemble. This implies that our approach can perform effective knowledge adaptation even not accessing the private dataset.

\subsection{Evaluating Face Verification on LFW}
In the second experiment, we extensively evaluate the performance of the proposed approach on face verification task under various input resolutions $\textbf{p}=\{96,64,32,16\}$, supervision signals $\textbf{x}=\{$c,s,dc,sc$\}$ and distilled teachers $\textbf{T}=\{$O,V,C,E$\}$. Here, the abbreviations of supervision signals stand for only class-level supervision  (c), cross-dataset distillation when discarding class-level supervision (s), direct distillation in addition to class-level supervision (dc) and cross-dataset distillation in addition to class-level supervision (sc), respectively. In the case of dc, the adaptation module is discarded and the student model directly learns to mimic the $2,048$ or $1,024$ dimensional pretrained features from the teacher. The abbreviations of distilled teachers represent no teacher (O), VGGFace2 (V), CosFace (C) and their ensemble (E), respectively.
In this context, a student model of a specific setting is represented as the combination of its resolution, supervision signal and distilled teacher, \eg, S-p-x-t where p$\in\textbf{p}$, x$\in\textbf{x}$ and t$\in\textbf{T}$. For conciseness we represent a teacher model with the same rule, \eg~T-112-sc-C represents the cross-dataset distilled CosFace teacher model at $112 \times 112$ resolution and adapted with the supervision signal sc. We also use T-224-o-V and T-112-o-C to represent the original VGGFace2 and CosFace teacher models without adaptation, respectively.

From Fig. \ref{fig:res-lfw}, several important observations can be summarized.
First, the accuracy gradually degenerates as the resolution decreases, which is as expected. Although S-96-sc-$\{$V,C,E$\}$ still perform worse than their teachers, they reach a reasonably good accuracy (\eg,$92.67\%$ with S-96-sc-E) with a tiny model size of $0.21$M parameters, indicating that bridge distillation provides a way to redeploy existing heavy models on resource-limited devices.
Second, the student model with only class-level supervision S-p-c-O perform consistently worse than S-p-sc-$\{$V,C,E$\}$, where p$\in\textbf{p}$. This implies that the learned model without cross-dataset distillation may lack the capability of capturing discriminative high-resolution details due to direct training on low-resolution faces. On the contrary, the student models can explicitly learn to reconstruct the missing details by mimicking the high-resolution teacher's knowledge.
Third, the student model distilled from the ensemble outperforms its corresponding model distilled from single teacher, revealing that transferring more informative knowledge will result in better performance improvement.
Finally, to show the impact of cross-dataset distillation, we compare each S-16-sc-t with its direct distillation model S-16-dc-t, where t$\in\{V,C\}$. The results show that combining adaptation via cross-dataset distillation can boost the inference performance. We further note that such improvement is consistent across various resolutions and different distilled teachers. Therefore, it is fair to conclude that the adaptive nature of the proposed bridge distillation approach is necessary to learn the discriminative characteristics. We suspect that introducing the distillation loss to complement the classification loss can prevent the training bias towards target domain much, leading to less overfitting.

Moreover, the detail is shown in the Receiver Operating characteristics Curves (ROC) on LFW (see Fig. \ref{fig:roc-lfw}). Compared with the other five models, our S-16-sc-E model achieves the best performance. In particular, when the false positive rate is 10.0\%, the true positive rate of S-16-sc-E is 80.4\%, which is 9.6\% higher than that of S-16-c-O, indicating the effectiveness of our proposed bridge distillation approach under low false positive rate.

\begin{figure}[t]
	\centering{\includegraphics[width=1.0\linewidth]{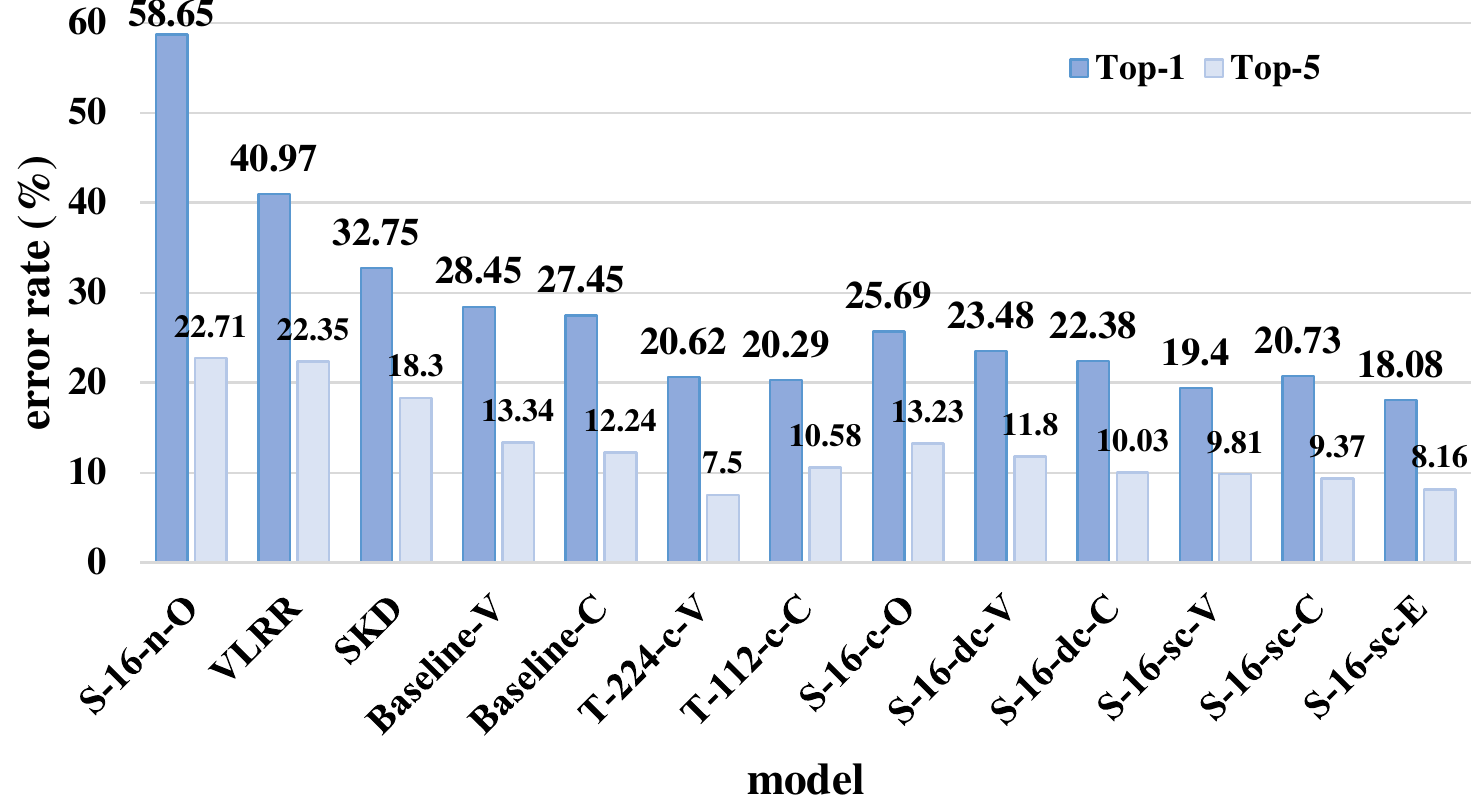}}
	\caption{The performance of various models in evaluating face identification on UCCS.}
	\label{fig:res-uccs}
\end{figure}

\subsection{Evaluating Face Identification on UCCS}
In the final experiment, we make comparisons with the state-of-the-art model on the low-resolution face identification task using the challenging UCCS dataset. On this dataset, we directly compare with the Very Low-Resolution Recognition (VLRR) \cite{wang2016cvpr} and Selective Knowledge Distillation (SKD) \cite{Ge2019TIP}, which are two state-of-the-art low-resolution face recognition models that work at resolution $16 \times 16$.

We follow similar experimental settings, randomly dividing the $180$-subject subset into training set ($4,500$ images) and testing set ($935$ images), where face images are normalized to $16 \times 16$ and the identities in two sets are exclusive, fine-tuning the softmax layers of various student models S-16-\textbf{x}-\textbf{T} on the training set, and performing evaluation on the testing set. We also fine-tune the teacher models T-224-c-V and T-112-c-C without distillation on UCCS in a similar manner. Then, we directly distil them to two baselines ({Baseline-V} and {Baseline-C}). In addition, we train a new model S-16-n-O on UCCS from scratch for comparison.

Fig.\ref{fig:res-uccs} shows the results, where top-1 and top-5 error rates are reported. We can find that the fine-tuned teachers T-224-c-V and T-112-c-C as well as the distilled baselines all perform significantly better than VLRR and SKD but S-16-n-O is worse, implying that the models pretrained on external datasets can provide valuable prior knowledge on low-resolution recognition problem. Therefore, all the fine-tuned students give much lower error rates than VLRR and SKD while they have greatly reduced parameters comparing with the fine-tuned teachers. Moreover, the fine-tuned student models with cross-dataset distillation achieve best results. We suspect that adapting the rich pretrained knowledge from the well-learned teacher allows the student to successfully reconstruct informative high-resolution details, even when the input resolution is very low.

\begin{table}[t]
   \centering
   \caption{The resource costs of various models. Our model costs much less parameters and works at very low resolution.}
   \begin{tabular}{ccccc}
     \hline
     Model                            &Resolution            &\#Para. &FLOPs &Year\\
     \hline
     DeepID \cite{Sun2014CVPR}        &39$\times$31    &17M     &352M  &2014\\
     DeepID2 \cite{Sun2014NIPS}       &55$\times$47    &10M     &194M  &2014\\
     MobileID \cite{Luo2016AAAI}      &55$\times$47    &2M      &98M   &2016\\
     ShiftFaceNet \cite{wu2018cvpr}   &224$\times$224  &0.78M   &344M  &2018\\
     LRFRW \cite{li2019low}          &64$\times$64    &4.2M    &159M  &2019\\
     SKD \cite{Ge2019TIP}             &16$\times$16    &0.79M   &2.43M &2019\\
     OUR                              &16$\times$16    &0.11M   &1.51M &-\\
     \hline
   \end{tabular}
   \label{Table:res-lfw}
 \end{table}

\begin{table}[t]\small
  \caption{Inference memory (MB) and speed (faces per second). GPU:
  Nvidia K80, CPU: Intel 2.6GHZ, Mobile: Snapdragon 821}
  \centering{
  \begin{tabular}{cccccc}
  \toprule
  \multirow{2}*{Model}  &\multirow{2}*{Memory}  &\multicolumn{3}{c}{Speed} \\
  \cline{3-5}
                        &                       &GPU     &CPU   &Mobile     \\
  \midrule
  T-224-o-V             &35.41                  &27      &0.01  &1.72       \\
  T-112-o-C             &18.86                  &483     &5     &3.6        \\
  S-96-sc-T             &2.005                  &2,941   &33    &59         \\
  S-64-sc-T             &0.892                  &5,000   &81    &121        \\
  S-32-sc-T             &0.224                  &12,987  &335   &206        \\
  S-16-sc-T             &0.057                  &14,705  &934   &763        \\
  \bottomrule
  \end{tabular}}
  \label{Table:inference-time}
  \end{table}

\subsection{Efficiency Analysis}
The student can mimic the teacher's performance while largely reducing the cost in storage, memory and computation.

As shown in Table \ref{Table:res-lfw}, we compare with state-of-the-art light-weight face models. DeepID \cite{Sun2014CVPR} and DeepID2 \cite{Sun2014NIPS} models take low resolution inputs and ensemble tens of networks, leading to large computation complexity. MobileID \cite{Luo2016AAAI} compresses the DeepID2 model with fewer parameters of 2M, which still needs 98M FLOPs. Recently, ShiftFaceNet \cite{wu2018cvpr} applies a light-weight network with 0.79M parameters to recognize $224\times224$ face images, thus still costs heavy commutation burden. LRFRW \cite{li2019low} takes a low resolution input of $64\times64$ and 4.2M parameters, but the FLOPs reach 159M. The most recent SKD \cite{Ge2019TIP} can recognize $16\times16$ face images while costing 0.79M parameters. Compared with them, our student model only has much less parameters of 0.11M and cost less computations, while working at very low resolution.

From Table \ref{Table:inference-time}, we can find that the memory has a reduction factor of $18\times$, $40\times$, $158\times$ and $621\times$ for the students at resolution $96 \times 96$, $64 \times 64$, $32 \times 32$ and $16\times16$, respectively, compared with VGGFace2 teacher, or $9\times$, $21\times$, $84\times$ and $331\times$ when compared with CosFace teacher. In particular, the memory is only $0.057$MB for the $16\times16$ student.
Beyond the significant saving in memory, the computational cost still greatly reduces either. As shown in Table \ref{Table:inference-time}, on a NVIDIA K80 GPU, while the inference speed is only $27$ or $483$ faces per second with the teacher model VGGFace2 or CosFace, it reaches up to $2,941$, $5,000$, $12,987$ and $14,705$ with the S-p-sc-t architecture, where t=$\{$V,C,E$\}$ and the resolution sizes p are 96, 64, 32 and 16, respectively. Even in CPU, the inference speed is also remarkably fast. In particular, when the models are deployed on mobile devices, the inference time is only $1.31$ms for the model S-16-sc-T. As a result, the proposed tiny student model S-16-sc-T is able to process $14,705$, $934$ or $763$ faces per second on a GPU, CPU or a mobile phone. These results indicate that our bridge distillation approach provides a practical solution to redeploy existing heavy pretrained models on low-end devices.

\subsection{Comparison Analysis}

\begin{table}[t]
  \caption{Face recognition performance (\%) with naive hallucination-based approaches. Here, a model is represented as \emph{r-m} where \emph{r} and \emph{m} denote the resolution of input faces and hallucination method. BI: Bilinear interpolation, SR: Super Resolution.}
  \centering{
  \begin{tabular}{cccccc}
  \toprule
  \multirow{2}*{Recognizer}             &\multicolumn{2}{c}{LFW} & &\multicolumn{2}{c}{UCCS (Top1/Top5)} \\
  \cline{2-3} \cline{5-6}
                                   &16-BI    &16-SR        &  &16-BI          &16-SR     \\
  \midrule
  CenterLoss \cite{Wen2016ECCV}    &75.62    &84.13        &  &77.75/91.13  &84.23/94.51       \\
  SphereFace \cite{liu2017cvpr}    &76.62    &85.62        &  &78.73/86.20  &80.85/87.75        \\
  CosFace \cite{wang2018cosface}   &84.33    &86.07        &  &91.83/96.76  &94.08/97.61         \\
  VGGFace2 \cite{cao2018fg}        &87.95    &88.55        &  &84.65/92.68  &86.34/95.35        \\
  ArcFace \cite{deng2019cvpr}      &86.30    &86.97        &  &88.73/95.77  &90.56/96.48        \\
  \midrule
  Average                          &82.16    &86.27        &  &84.34/92.51  &87.21/94.34        \\
  \bottomrule
  \end{tabular}}
  \label{Table:res-hallucination}
  \end{table}

\begin{table}[t]
   \centering
   \caption{Recognition accuracy (\%) on LFW benchmark with different face recognition losses in classification task.}
   \begin{tabular}{ccccc}
     \toprule
     Loss                             &S-16-dc-V   &S-16-sc-V  &S-16-sc-C  &S-16-sc-E  \\
     \midrule
     CenterLoss \cite{Wen2016ECCV}    &80.90       &84.37      &85.12      &85.15\\
     SphereFace+ \cite{liu2018nips}   &83.03       &86.20      &86.33      &86.22\\
     CosFace \cite{wang2018cosface}   &82.67       &84.33      &84.50      &84.75\\
     ArcFace \cite{deng2019cvpr}      &82.88       &85.62      &85.00      &85.73\\
     \midrule
     Average                          &82.37       &85.13      &85.24      &85.46\\
     \midrule
     Baseline(Softmax)                &82.38       &83.95      &85.50      &85.88\\
     \bottomrule
   \end{tabular}
   \label{Table:res-lfw-loss}
 \end{table}

To further study the effectiveness of the proposed approach, we conduct two experimental comparisons, including the comparison with naive hallucination-based approaches and the models with different face recognition losses.

First, we conduct the experimental comparison on naive hallucination-based approaches on LFW and UCCS benchmarks. In our experiment, we check two hallucination methods, BI (bilinear interpolation) and SR (a recent FSRNet approach \cite{chen2018cvpr}), as well as five high-resolution face recognition models (CenterLoss \cite{Wen2016ECCV}, SphereFace \cite{liu2017cvpr}, CosFace \cite{wang2018cosface}, VGGFace2 \cite{cao2018fg} and ArcFace \cite{deng2019cvpr}). Note that these models are provided by their authors with parameters pretrained on specific datasets. The input face resolution is $16\times16$. Tab. \ref{Table:res-hallucination} shows the results.
From the results, we can find that the hallucination-based approaches generally can achieve good recognition accuracy. For example, the average recognition accuracy can reach $82.16\%$ on LFW even using simply bilinear interpolation to hallucinate face images. It is also noted that complex hallucination method often leads to larger performance improvement while costing more computational burden in the reconstruction process. Moreover, SphereFace performs worse than CosFace and ArcFace on UCCS, since it is pretrained on CASIA-WebFace dataset \cite{yi2014webface} where the face images have very different distribution from the surveillance faces in UCCS. Our approach achieves comparable performance while costing much less computational and memory resources. For example, our S-16-sc-E student model gives an accuracy of $85.88\%$ on LFW and $91.84\%$@Top5 on UCCS.

We next check the impact of different loss functions for classification task on recognition performance. Beyond the baseline softmax loss, we conduct the experiments with four kinds of face recognition losses (CenterLoss \cite{Wen2016ECCV}, SphereFace+ \cite{liu2018nips}, CosFace \cite{wang2018cosface} and ArcFace \cite{deng2019cvpr}) on four student models. Tab. \ref{Table:res-lfw-loss} shows the results, where our proposed approach has consistent performance boost under various kinds of face recognition losses when having better distillation losses. It implies that our proposed approach is loss-agnostic.

\section{Conclusion}
This paper proposes a novel bridge distillation approach to solve low-resolution face recognition tasks with limited resources. The core of this approach is the efficient teacher-student framework that relies on novel cross-dataset distillation and resolution-adapted distillation algorithms, which first adapt the teacher model to preserve the discriminative high-resolution details and then use them to supervise the training of the student models. Extensive experimental results show that the proposed approach is able to transfer informative high-resolution knowledge from the teacher to the student, leading to significantly reduced model with much fewer parameters and extremely fast inference speed. In the future work, we will explore the possibility of multi-bridge knowledge distillation on extensive visual tasks.

\myPara{Acknowledgement}. This work was partially supported by grants from the National Natural Science Foundation of China (61772513 \& 61922006), the project from Beijing Municipal Science and Technology Commission (Z191100007119002), Beijing Nova Program (Z181100006218063), and Beijing Natural Science Foundation (19L2040). Shiming Ge is also supported by the Youth Innovation Promotion Association, Chinese Academy of Sciences.

\bibliographystyle{IEEEtran}
\bibliography{bibBD}

\begin{thebibliography}{10}
\providecommand{\url}[1]{#1}
\csname url@samestyle\endcsname
\providecommand{\newblock}{\relax}
\providecommand{\bibinfo}[2]{#2}
\providecommand{\BIBentrySTDinterwordspacing}{\spaceskip=0pt\relax}
\providecommand{\BIBentryALTinterwordstretchfactor}{4}
\providecommand{\BIBentryALTinterwordspacing}{\spaceskip=\fontdimen2\font plus
\BIBentryALTinterwordstretchfactor\fontdimen3\font minus \fontdimen4\font\relax}
\providecommand{\BIBforeignlanguage}[2]{{%
\expandafter\ifx\csname l@#1\endcsname\relax
\typeout{** WARNING: IEEEtran.bst: No hyphenation pattern has been}%
\typeout{** loaded for the language `#1'. Using the pattern for}%
\typeout{** the default language instead.}%
\else
\language=\csname l@#1\endcsname
\fi
#2}}
\providecommand{\BIBdecl}{\relax}
\BIBdecl

\bibitem{Ge2019TIP}
S.~Ge, S.~Zhao, C.~Li, and J.~Li, ``Low-resolution face recognition in the wild via selective knowledge distillation,'' \emph{IEEE Transactions on Image Processing}, vol.~28, no.~4, pp. 2051--2062, 2019.

\bibitem{Schroff2015CVPR}
F.~Schroff, D.~Kalenichenko, and J.~Philbin, ``{FaceNet}: A unified embedding for face recognition and clustering,'' in \emph{IEEE Conference on Computer Vision and Pattern Recognition (CVPR)}, 2015, pp. 815--823.

\bibitem{liu2017cvpr}
W.~Liu, Y.~Wen, Z.~Yu \emph{et~al.}, ``Sphereface: Deep hypersphere embedding for face recognition,'' in \emph{IEEE Conference on Computer Vision and Pattern Recognition (CVPR)}, 2017, pp. 6738--6746.

\bibitem{zheng2018ring}
Y.~Zheng, D.~K. Pal, and M.~Savvides, ``Ring loss: Convex feature normalization for face recognition,'' in \emph{IEEE Conference on Computer Vision and Pattern Recognition (CVPR)}, 2018, pp. 5089--5097.

\bibitem{wang2018cosface}
H.~Wang, Y.~Wang, Z.~Zhou \emph{et~al.}, ``Cosface: Large margin cosine loss for deep face recognition,'' in \emph{IEEE Conference on Computer Vision and Pattern Recognition (CVPR)}, 2018, pp. 4510--4520.

\bibitem{cao2018fg}
Q.~Cao, L.~Shen, W.~Xie, O.~M. Parkhi, and A.~Zisserman, ``{VGGFace2}: {A} dataset for recognising faces across pose and age,'' in \emph{IEEE International Conference on Automatic Face and Gesture Recognition}, 2018, pp. 67--74.

\bibitem{wang2016cvpr}
Z.~Wang, S.~Chang, Y.~Yang, D.~Liu, and T.~Huang, ``Studying very low resolution recognition using deep networks,'' in \emph{IEEE Conference on Computer Vision and Pattern Recognition (CVPR)}, 2016, pp. 4792--4800.

\bibitem{Cheng2018AAAI}
B.~Cheng, D.~Liu, Z.~Wang \emph{et~al.}, ``Visual recognition in very low-quality settings: Delving into the power of pre-training,'' in \emph{National Conference on Artificial Intelligence (AAAI)}, 2018, pp. 8065--8066.

\bibitem{jian2015simultaneous}
M.~Jian and K.-M. Lam, ``Simultaneous hallucination and recognition of low-resolution faces based on singular value decomposition,'' \emph{IEEE Transactions on Circuits and Systems for Video Technology}, vol.~25, no.~11, pp. 1761--1772, 2015.

\bibitem{yang2015recognition}
M.-C. Yang, C.-P. Wei, Y.-R. Yeh, and Y.-C.~F. Wang, ``Recognition at a long distance: Very low resolution face recognition and hallucination,'' in \emph{International Conference on Biometrics}, 2015, pp. 237--242.

\bibitem{yu2017cvpr}
X.~Yu and F.~Porikli, ``Hallucinating very low-resolution unaligned and noisy face images by transformative discriminative autoencoders,'' in \emph{IEEE Conference on Computer Vision and Pattern Recognition (CVPR)}, 2017, pp. 5367--5375.

\bibitem{cheng2018accv}
Z.~Cheng, X.~Zhu, and S.~Gong, ``Low-resolution face recognition,'' in \emph{Asian Conference on Computer Vision (ACCV)}, 2018, pp. 605--621.

\bibitem{kolouri2015transport}
S.~Kolouri and G.~K. Rohde, ``Transport-based single frame super resolution of very low resolution face images,'' in \emph{IEEE Conference on Computer Vision and Pattern Recognition (CVPR)}, 2015, pp. 4876--4884.

\bibitem{biswas2012multidimensional}
S.~Biswas, K.~W. Bowyer, and P.~J. Flynn, ``Multidimensional scaling for matching low-resolution face images,'' \emph{IEEE Transactions on Pattern Analysis and Machine Intelligence}, vol.~34, no.~10, pp. 2019--2030, 2012.

\bibitem{ren2012coupled}
C.-X. Ren, D.-Q. Dai, and H.~Yan, ``Coupled kernel embedding for low-resolution face image recognition,'' \emph{IEEE Transactions on Image Processing}, vol.~21, no.~8, pp. 3770--3783, 2012.

\bibitem{Hinton2014NIPSW}
G.~Hinton, J.~Dean, and O.~Vinyals, ``Distilling the knowledge in a neural network,'' in \emph{Workshop on Advances in Neural Information Processing Systems}, 2014, pp. 1--9.

\bibitem{Romero2015ICLR}
\BIBentryALTinterwordspacing
A.~Romero, N.~Ballas, S.~E. Kahou \emph{et~al.}, ``{FitNets}: Hints for thin deep nets,'' in \emph{International Conference on Learning Representations (ICLR)}, 2015. [Online]. Available: \url{http://arxiv.org/abs/1412.6550}
\BIBentrySTDinterwordspacing

\bibitem{Luo2016AAAI}
P.~Luo, Z.~Zhu, Z.~Liu \emph{et~al.}, ``Face model compression by distilling knowledge from neurons,'' in \emph{Proceedings of the AAAI Conference on Artificial Intelligence (AAAI)}, 2016, pp. 3560--3566.

\bibitem{lopezpaz2016iclr}
D.~Lopezpaz, L.~Bottou, B.~Scholkopf, and V.~Vapnik, ``Unifying distillation and privileged information,'' in \emph{International Conference on Learning Representations (ICLR)}, 2016.

\bibitem{radosavovic2018cvpr}
I.~Radosavovic, P.~Dollar, R.~Girshick \emph{et~al.}, ``Data distillation: Towards omni-supervised learning,'' in \emph{IEEE Conference on Computer Vision and Pattern Recognition (CVPR)}, 2018, pp. 4119--4128.

\bibitem{Phuong2019icml}
M.~Phuong and C.~Lampert, ``Towards understanding knowledge distillation,'' in \emph{International Conference on Machine Learning (ICML)}, 2019, pp. 5142--5151.

\bibitem{Sun2014NIPS}
Y.~Sun, X.~Wang, and X.~Tang, ``Deep learning face representation by joint identification-verification,'' \emph{Advances in Neural Information Processing Systems}, pp. 1988--1996, 2014.

\bibitem{liu2016icml}
W.~Liu, Y.~Wen, Z.~Yu, and M.~Yang, ``Large-margin softmax loss for convolutional neural networks,'' in \emph{Proceedings of International Conference on Machine Learning (ICML)}, 2016, pp. 507--516.

\bibitem{Zhang2017RangeLoss}
X.~Zhang, Z.~Fang, Y.~Wen \emph{et~al.}, ``Range loss for deep face recognition with long-tail,'' in \emph{IEEE International Conference on Computer Vision (ICCV)}, 2017, pp. 5409--5418.

\bibitem{ledig2016photo}
C.~Ledig, L.~Theis, F.~Husz¨¢r \emph{et~al.}, ``Photo-realistic single image super-resolution using a generative adversarial network,'' in \emph{IEEE Conference on Computer Vision and Pattern Recognition (CVPR)}, 2017, pp. 105--114.

\bibitem{Zhang2018ECCV}
K.~Zhang, Z.~Zhang, C.-W. Cheng, W.~H. Hsu, Y.~Qiao, W.~Liu, and T.~Zhang, ``Super-identity convolutional neural network for face hallucination,'' in \emph{European Conference on Computer Vision (ECCV)}, 2018, pp. 196--211.

\bibitem{li2019low}
P.~Li, L.~Prieto, D.~Mery, and P.~J. Flynn, ``On low-resolution face recognition in the wild: Comparisons and new techniques,'' \emph{IEEE Transactions on Information Forensics and Security}, vol.~14, no.~8, pp. 2000--2012, 2019.

\bibitem{liu2017nips}
W.~Liu, Y.~Zhang, X.~Li, Z.~Liu, B.~Dai, T.~Zhao, and L.~Song, ``Deep hyperspherical learning,'' in \emph{Advances in Neural Information Processing Systems}, 2017, pp. 3950--3960.

\bibitem{wang2018additive}
F.~Wang, J.~Cheng, W.~Liu, and H.~Liu, ``Additive margin softmax for face verification,'' \emph{IEEE Signal Processing Letters}, vol.~25, no.~7, pp. 926--930, 2018.

\bibitem{liu2018nips}
W.~Liu, R.~Lin, Z.~Liu, L.~Liu, Z.~Yu, B.~Dai, and L.~Song, ``Learning towards minimum hyperspherical energy,'' in \emph{Advances in Neural Information Processing Systems}, 2018, pp. 6225--6236.

\bibitem{deng2019cvpr}
J.~Deng, J.~Guo, and S.~Zafeiriou, ``Arcface: Additive angular margin loss for deep face recognition,'' in \emph{IEEE Conference on Computer Vision and Pattern Recognition (CVPR)}, 2019, pp. 4690--4699.

\bibitem{jiang2016cdmma}
J.~Jiang, R.~Hu, Z.~Wang, and Z.~Cai, ``{CDMMA}: Coupled discriminant multi-manifold analysis for matching low-resolution face images,'' \emph{Signal Processing}, vol. 124, pp. 162--172, 2016.

\bibitem{wang2016pose}
X.~Wang, H.~Hu, and J.~Gu, ``Pose robust low-resolution face recognition via coupled kernel-based enhanced discriminant analysis,'' \emph{IEEE/CAA Journal of Automatica Sinica}, vol.~3, no.~2, pp. 203--212, 2016.

\bibitem{haghighat2017low}
M.~Haghighat and M.~Abdel-Mottaleb, ``Low resolution face recognition in surveillance systems using discriminant correlation analysis,'' in \emph{IEEE International Conference on Automatic Face and Gesture Recognition}, 2017, pp. 912--917.

\bibitem{mudunuri2016low}
S.~P. Mudunuri and S.~Biswas, ``Low resolution face recognition across variations in pose and illumination,'' \emph{IEEE Transactions on Pattern Analysis and Machine Intelligence}, vol.~38, no.~5, pp. 1034--1040, 2016.

\bibitem{shekhar2017synthesis}
S.~Shekhar, V.~M. Patel, and R.~Chellappa, ``Synthesis-based robust low resolution face recognition,'' \emph{arXiv}, 2017.

\bibitem{li2015multi}
X.~Li, W.-S. Zheng, X.~Wang \emph{et~al.}, ``Multi-scale learning for low-resolution person re-identification,'' in \emph{IEEE International Conference on Computer Vision (ICCV)}, 2015, pp. 3765--3773.

\bibitem{pong2014multi}
K.-H. Pong and K.-M. Lam, ``Multi-resolution feature fusion for face recognition,'' \emph{Pattern Recognition}, vol.~47, no.~2, pp. 556--567, 2014.

\bibitem{herrmann2016low}
C.~Herrmann, D.~Willersinn, and J.~Beyerer, ``Low-resolution convolutional neural networks for video face recognition,'' in \emph{IEEE International Conference on Advanced Video and Signal-based Surveillance (AVSS)}, 2016, pp. 221--227.

\bibitem{rozantsev2018cvpr}
A.~Rozantsev, M.~Salzmann, and P.~Fua, ``Residual parameter transfer for deep domain adaptation,'' in \emph{IEEE Conference on Computer Vision and Pattern Recognition (CVPR)}, 2018, pp. 4339--4348.

\bibitem{chen2017nips}
G.~Chen, W.~Choi, X.~Yu \emph{et~al.}, ``Learning efficient object detection models with knowledge distillation,'' in \emph{Advances in Neural Information Processing Systems}, 2017, pp. 742--751.

\bibitem{yim2017gift}
J.~Yim, D.~Joo, J.~Bae, and J.~Kim, ``A gift from knowledge distillation: Fast optimization, network minimization and transfer learning,'' in \emph{IEEE Conference on Computer Vision and Pattern Recognition (CVPR)}, 2017, pp. 7130--7138.

\bibitem{su2017bmvc}
J.-C. Su and S.~Maji, ``Adapting models to signal degradation using distillation,'' in \emph{British Machine Vision Conference (BMVC)}, 2017.

\bibitem{rebuffi2017cvpr}
S.~Rebuffi, A.~Kolesnikov, G.~Sperl, and C.~H. Lampert, ``{iCaRL}: Incremental classifier and representation learning,'' in \emph{IEEE Conference on Computer Vision and Pattern Recognition (CVPR)}, 2017, pp. 5533--5542.

\bibitem{li2018pami}
Z.~Li and D.~Hoiem, ``Learning without forgetting,'' \emph{IEEE Transactions on Pattern Analysis and Machine Intelligence}, vol.~40, no.~12, pp. 2935--2947, 2018.

\bibitem{busto2017open}
P.~P. Busto and J.~Gall, ``Open set domain adaptation,'' in \emph{IEEE International Conference on Computer Vision (ICCV)}, 2017, pp. 754--763.

\bibitem{Parkhi2015BMVC}
O.~M. Parkhi, A.~Vedaldi, and A.~Zisserman, ``Deep face recognition,'' in \emph{British Machine Vision Conference (BMVC)}, vol.~1, no.~3, 2015, p.~6.

\bibitem{yi2014webface}
D.~Yi, Z.~Lei, S.~Liao, and S.~Z. Li, ``Learning face representation from scratch,'' \emph{arXiv}, 2014.

\bibitem{Redmon2017CVPR}
J.~Redmon and A.~Farhadi, ``Yolo9000: Better, faster, stronger,'' in \emph{IEEE Conference on Computer Vision and Pattern Recognition (CVPR)}, 2017, pp. 7263--7271.

\bibitem{He2016CVPR}
K.~He, X.~Zhang, S.~Ren, and J.~Sun, ``Deep residual learning for image recognition,'' in \emph{IEEE Conference on Computer Vision and Pattern Recognition (CVPR)}, 2016, pp. 770--778.

\bibitem{Bansal2016UMDFaces}
A.~Bansal, A.~Nanduri, C.~Castillo \emph{et~al.}, ``Umdfaces: An annotated face dataset for training deep networks,'' in \emph{IEEE International Joint Conference on Biometrics (IJCB)}, 2017, pp. 464--473.

\bibitem{Learned2016Labeled}
E.~Learned-Miller, G.~B. Huang, A.~RoyChowdhury \emph{et~al.}, ``Labeled faces in the wild: A survey,'' \emph{Advances in Face Detection and Facial Image Analysis}, 2016.

\bibitem{Sapkota2013BTAS}
A.~Sapkota and T.~E. Boult, ``Large scale unconstrained open set face database,'' in \emph{International Conference on Biometrics}, 2014, pp. 1--8.

\bibitem{Krizhevsky2012NIPS}
A.~Krizhevsky, I.~Sutskever, and G.~E. Hinton, ``Imagenet classification with deep convolutional neural networks,'' in \emph{Advances in Neural Information Processing Systems}, 2012, pp. 1097--1105.

\bibitem{Abadi2016OSDI}
M.~Abadi, A.~Agarwal, P.~Barham \emph{et~al.}, ``Tensorflow: A system for large-scale machine learning,'' in \emph{USENIX Symposium on Operating Systems Design and Implementation (OSDI)}, vol.~16, 2016, pp. 265--283.

\bibitem{Sun2014CVPR}
Y.~Sun, X.~Wang, and X.~Tang, ``Deep learning face representation from predicting 10,000 classes,'' in \emph{IEEE Conference on Computer Vision and Pattern Recognition (CVPR)}, 2014, pp. 1891--1898.

\bibitem{wu2018cvpr}
B.~Wu, A.~Wan, X.~Yue \emph{et~al.}, ``Shift: A zero flop, zero parameter alternative to spatial convolutions,'' in \emph{IEEE Conference on Computer Vision and Pattern Recognition (CVPR)}, 2018, pp. 9127--9135.

\bibitem{Wen2016ECCV}
W.~Wu, M.~Kan, X.~Liu \emph{et~al.}, ``A discriminative feature learning approach for deep face recognition,'' in \emph{European Conference on Computer Vision (ECCV)}, 2016, pp. 499--515.

\bibitem{chen2018cvpr}
Y.~Chen, Y.~Tai, X.~Liu, C.~Shen, and J.~Yang, ``Fsrnet: End-to-end learning face super-resolution with facial priors,'' in \emph{IEEE Conference on Computer Vision and Pattern Recognition (CVPR)}, 2018, pp. 2492--2501.

\end{thebibliography}

%




%
\end{document}